\documentclass[sigconf]{acmart}
\AtBeginDocument{%
  \providecommand\BibTeX{{%
    \normalfont B\kern-0.5em{\scshape i\kern-0.25em b}\kern-0.8em\TeX}}}

\setcopyright{acmcopyright}
\copyrightyear{2020}
\acmYear{2020}
\acmDOI{10.1145/1122445.1122456}

\acmConference[ICONS '20]{International Conference on Neuromorphic Systems '20: }{July 28--30, 2020}{Virtual Conference}
\acmPrice{15.00}
\acmISBN{978-1-4503-XXXX-X/18/06}

\begin{document}

\title{An Astrocyte-Modulated Neuromorphic Central Pattern Generator for Hexapod Robot Locomotion on Intel's Loihi}

\author{Ioannis Polykretis}
\affiliation{%
  \institution{Computational Brain Lab}
  \city{Rutgers University}
  \authornote{This work is supported by Intel's NRC grant award. IP is partially funded by the Onassis foundation scholarship. }
}


\author{Guangzhi Tang}
\affiliation{%
  \institution{Computational Brain Lab}
  \city{Rutgers University}
}

\author{Konstantinos P. Michmizos}
\affiliation{%
  \institution{Computational Brain Lab}
  \city{Rutgers University}
}
\email{konstantinos.michmizos@cs.rutgers.edu}

\renewcommand{\shortauthors}{Polykretis et al.}

\begin{abstract}
  Locomotion is a crucial challenge for legged robots that is addressed ``effortlessly'' by biological networks abundant in nature, named central pattern generators (CPG). The multitude of CPG network models that have so far become biomimetic robotic controllers is not applicable to the emerging neuromorphic hardware, depriving mobile robots of a robust walking mechanism that would result in inherently energy-efficient systems. Here, we propose a brain-morphic CPG controler based on a comprehensive spiking neural-astrocytic network that generates two gait patterns for a hexapod robot. Building on the recently identified astrocytic mechanisms for neuromodulation, our proposed CPG architecture is seamlessly integrated into Intel’s Loihi neuromorphic chip by leveraging a real-time interaction framework between the chip and the robotic operating system (ROS) environment, that we also propose. Here, we demonstrate that a Loihi-run CPG can be used to control a walking robot with robustness to sensory noise and varying speed profiles. Our results pave the way for scaling this and other approaches towards Loihi-controlled locomotion in autonomous mobile robots.

\end{abstract}

\begin{CCSXML}
<ccs2012>
<concept>
<concept_id>10010147.10010341.10010370</concept_id>
<concept_desc>Computing methodologies~Simulation evaluation</concept_desc>
<concept_significance>500</concept_significance>
</concept>
<concept>
<concept_id>10010405.10010444.10010087.10010091</concept_id>
<concept_desc>Applied computing~Biological networks</concept_desc>
<concept_significance>500</concept_significance>
</concept>
<concept>
<concept_id>10010405.10010444.10010095</concept_id>
<concept_desc>Applied computing~Systems biology</concept_desc>
<concept_significance>500</concept_significance>
</concept>
</ccs2012>
\end{CCSXML}

\ccsdesc[500]{Computing methodologies~Simulation evaluation}
\ccsdesc[500]{Applied computing~Biological networks}
\ccsdesc[500]{Applied computing~Systems biology}

\keywords{Central Pattern Generators, Astrocytes, Hexapod robot, Loihi, Neuromorphic Controller, Locomotion}


\maketitle

\section{Introduction}
Mobile robots have become successful in various applications including manned and autonomous vehicles \cite{autonomous}, rescue missions \cite{rescue}, and planetary or underwater exploration \cite{underwater, planetary}. With wheeled robots being challenged by uneven terrains, an alternative is legged robots that mimic animal locomotion \cite{crespi2006, sartoretti2018}. For controlling their articulated limbs, robots often utilize central pattern generators (CPG), a long-debated control mechanism \cite{illis1995, bussel1996} for a variety of repetitive behaviors that are not restricted to locomotion \cite{guertin2009, grillner1975}, as it extends from respiration \cite{von1983} to mastication \cite{lund2006}. Functionally, CPGs are specialized neuronal networks that can generate periodic output in the absence of a periodic input. 

Dynamical systems that mimic the oscillatory activity of CPG networks are commonly used to control robotic locomotion as a repetitive behavior \cite{crespi2006, ijspeert2008, righetti2006}. Multiple CPGs are combined to endorse robots with a remarkable set of gait patterns \cite{buono2001, barron2010}, motion speed variation \cite{crespi2013, van2015} and terrain adaptation \cite{sartoretti2018}. Although such controllers are typically validated based on accuracy, robustness and velocity of the movement, mobile robots deployed in real-world environments have also been increasingly challenged by their power consumption.

Large-scale neuromorphic processors emerged with a central promise to alleviate the energy limitations in mobile robots \cite{davies2018loihi,merolla2014million,schemmel2010wafer,furber2014spinnaker}. Indeed, it was recently demonstrated how brain-dictated spiking neural networks (SNN) developed on Intel's Loihi, can greatly improve power consumption of wheeled robots, while being as accurate as state-of-the-art methods \cite{tang2018, Guangzhi}. For legged robots, integrating CPGs into such chips requires a bottom-up rethinking of the conventional algorithms, since the mathematical formalism governing the oscillators is not translatable to non-Von Neumann architectures.

Inspired by the underlying neuronal connectome \cite{buono2001, mccrea2007}, SNNs are developed for controlling multiple gait patterns \cite{still2006, barron2010, rostro2015} and adjusting the speed of mobile robots \cite{crespi2006}. This body of prior work spurred a recent interest on neuromorphic implementations of CPGs \cite{gutierrez2019neuropod, perez2013, donati2018}. However, the application of SNN-controlled CPGs in real-world tasks is staggered by their intrinsic limitations. First, SNN-based CPG controllers mostly operate in open-loop, which inhibits their ability to change their behavior in response to their environment \cite{gutierrez2019neuropod, donati2014} (but see also \cite{steingrube2010}). Most of them are primed to drive a single pattern and have rather limited abilities in switching between distinct behaviors, especially in response to sensory stimuli. Second, even when they include sensory feedback, its spike-based encoding is prone to noise. This is mitigated by engineered principles that often deviate from biology. For example, some spike-based CPGs encode continuous ranges of joint movement in limited discrete states \cite{gutierrez2019neuropod}, which decreases motion smoothness and fine limb control. Other solutions artificially impose inter-limb delays that limit flexibility in the robot and interpretability in the model. Interestingly, a universal attribute of all the above approaches is that they regard CPG behavior as an emergent property of exclusively neuronal networks.

Overturning our assumptions about neural mechanisms, mounting evidence suggests that it is not only neurons that participate in CPGs \cite{condamine2018, acton2015}. Astrocytes, the most abundant type of \textit{non}-neuronal cells, in addition to their effects on synaptic plasticity \cite{depitta2016, polykretis2019computational}, are now seen as “a primary source for generating neural activity” \cite{RN128}. Linked to the long suspected role in modulating neural oscillations \cite{RN12, RN15, polykretis2018}, an astrocytic mechanism that controls CPGs was recently found. Astrocytes are now known to measure the intensity of sensory stimulation and conditionally switch the neuronal firing mode from regular spiking to bursting \cite{morquette2015}. With bursting emerging in a periodic fashion, astrocytes may switch neurons to pacemakers which shape patterns of repetitive behavior \cite{mccrea2007, morquette2015}. Paradoxically, as astrocytes are the underappreciated cells in brain hypotheses, have not been incorporated in possible CPG mechanisms. 

\begin{figure}[!t]
\vspace{-5.2pt}
\centering
\includegraphics[scale=1.05]{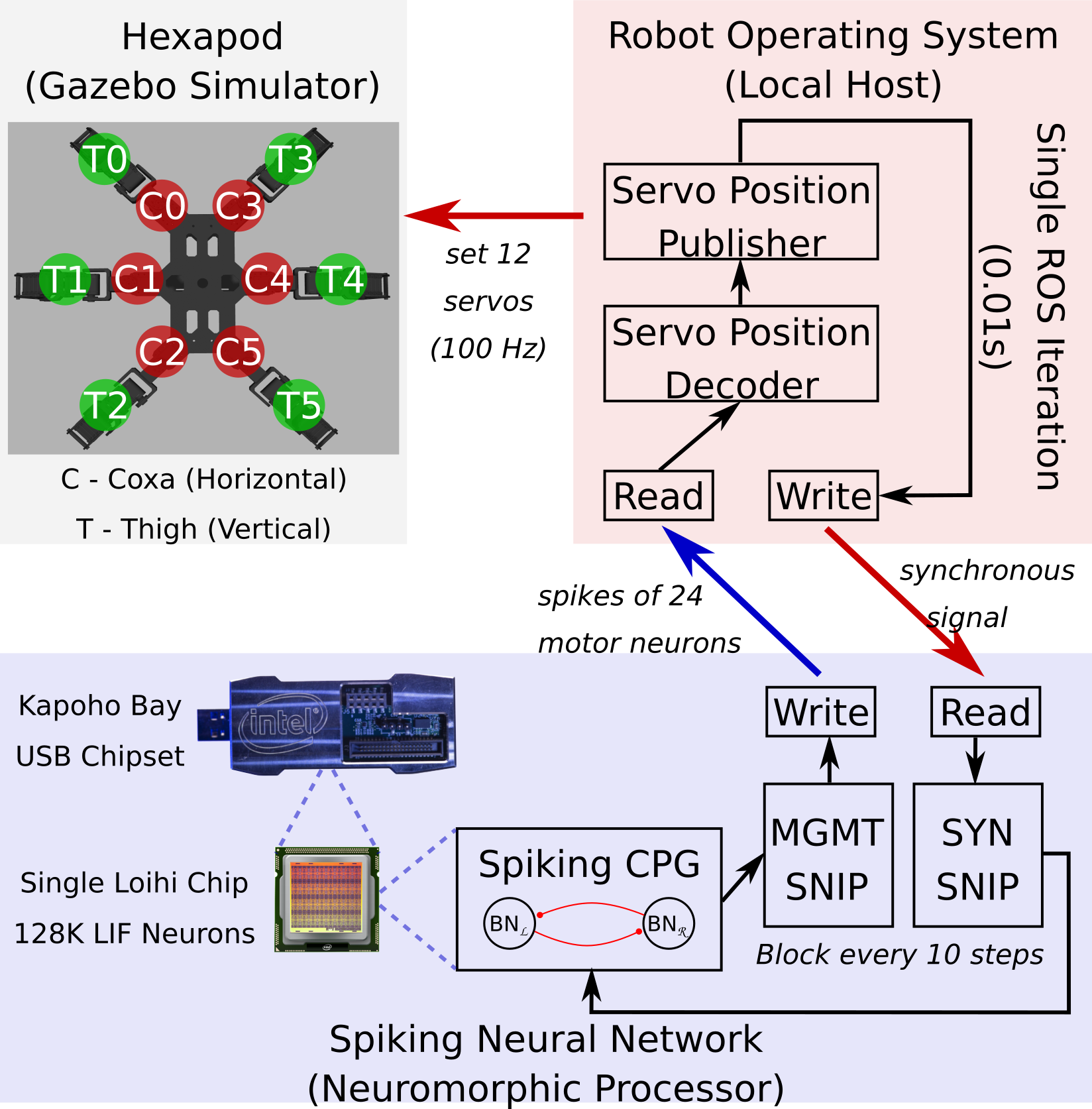}
\caption{A spiking CPG on Intel's Loihi neuromorphic processor controlling a hexapod in the Gazebo simulator. Loihi communicated with ROS in real-time using channel interfaces between the local host and the USB chipset. ROS decoded the spikes from 24 motor neurons on Loihi and controlled the positions of 12 servos supporting the hexapod.}
\vspace{-15.2pt}
\label{fig: overall}
\end{figure}

In this work, we developed a comprehensive neuronal-astrocytic CPG and implemented it on Loihi, Intel's neuromorphic chip, to control the locomotion of a hexapod robot. The astrocytic component i) modulated the bursting activity of the CPG neurons, ii) modified the behavior of the network by acting as a switch that turned the CPG on and off, and iii) filtered out the low-frequency input activity associated with sensory noise. Here, we also introduce a multi-spike joint control that results in a fine speed control. To achieve these results, we also developed the first interaction framework between Loihi and the robot operating system (ROS) that allows real-time control of robots. Overall, supported by the inherent energy-efficiency of neuromorphic approaches, our work aims to become one of the first hints that robotics will soon be, if they are not already are, the sweet spot for neuromorphic computing.

\section{Methods}
We took a bottom-up approach to design the CPG controller for the robotic hexapod. At the cellular level, we devised a bursting neuron by expanding the neuron model compartments that Loihi currently supports. We then utilized the neuron's bursting activity as a pacemaker, to control the flexion/extension of the joints. By connecting a number of such pacemaker neurons together, we created the CPG network, which drove the robot's walking pattern. To incorporate real-time control of the robotic locomotion, we designed an interactive framework between our Loihi-run CPG and the ROS environment. These steps are presented in more detail below.

\subsection{Design of a Bursting Neuron on Loihi}
Neurons that exhibited bursting activity were the building block of our spiking CPG network. However, spiking compartments on Loihi only support the leaky integrate-and-fire (LIF) neuronal model without any bursting dynamics, defined by the following equations: 
\begin{equation}
\begin{split}
    u_{i}(t) = u_{i}(t-1)\cdot dec_u + \sum_{j}^{} w_{ij}\cdot s_{j} 
\end{split}
\end{equation}
\begin{equation}
\begin{split}
    v_{i}(t) = v_{i}(t-1)\cdot dec_v + u_i(t)
\end{split}
\end{equation}
where $t$ is the Loihi step, $u_i$ is the current, $v_i$ is the voltage, $dec_u$ is the current decay, $dec_v$ is the voltage decay and $w_{ij}$ are the connection weights between the presynaptic compartment $j$ and the postsynaptic compartment $i$. 

We took advantage of Loihi's fully customizable multi-compart-mental neurons to introduce bursting dynamics to our neurons. Specifically, Loihi neurons are defined as binary trees, where either the spikes or the voltage of the child node are passed to the parent node. When a child node passes its voltage to a parent node, the voltage equation of the parent becomes:
\begin{equation}
\begin{split}
    v_{i}(t) = v_{i}(t-1)\cdot dec_v + u_i(t) + v_{ch}(t),
\end{split}
\end{equation}
where $v_{ch}(t)$ is the voltage of the compartment that is the child node to the parent compartment $i$. A child node can also pass its spikes to its parent node, which can be used as binary control signals. Another Loihi's feature that was crucial for our CPG implementation was its support of non-spiking compartments that had no voltage reset. We used those non-spiking compartments to inform their parent nodes about a voltage in a child node exceeding a threshold, without transmitting spikes. This was important to build a bursting neuron on the neuromorphic chip, as described below.  

\begin{figure}
\vspace{-5.2pt}
\centering
\includegraphics[scale=1.0]{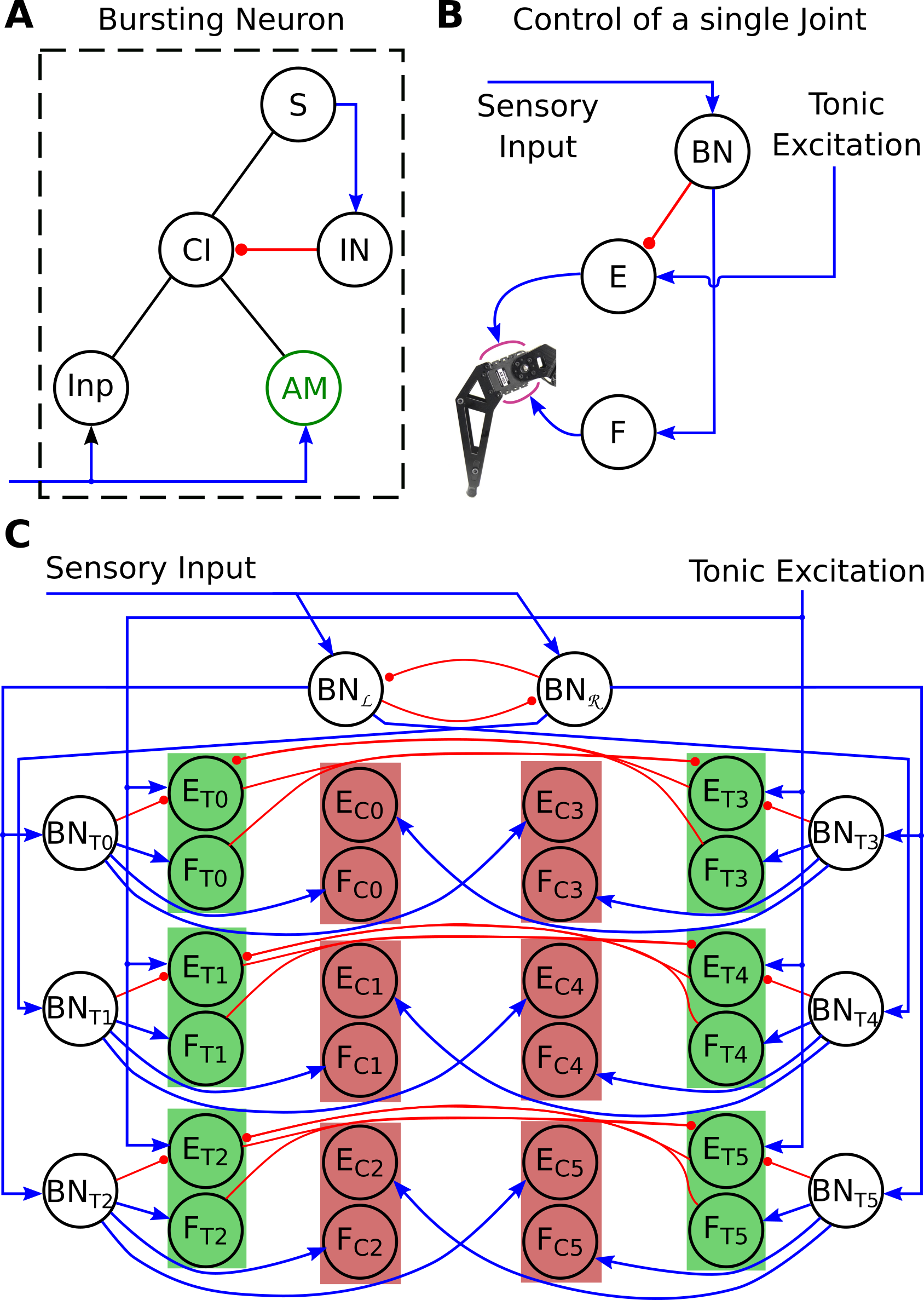}
\caption{Bottom-up architecture of the neuronal-astrocytic CPG network. A. Compartmental model of an astrocyte-dependent, conditionally bursting neuron on Loihi; B. Network controlling a single tibia joint of the robot; C. CPG architecture for the robot's locomotion. Dots denote connections. Blue arrows with arrowheads represent excitatory connections and red arrows with dotted heads show inhibitory connections.}
\label{fig: SNN}
\vspace{-15.2pt}
\end{figure}

This allowed us to design a conditionally bursting neuron as a multi-compartment neuron on Loihi (Fig. 2A), as described below.  The neuron's input drove two non-spiking compartments: a) the input compartment (Inp) that directly transformed the input current to voltage, and b) the astrocytic membrane compartment (AM) that integrated the current. The Conditional Integrator (CI) was Inp's and AM's non-spiking parent node. Inp passed its voltage to CI, while AM passed the information about exceeding its voltage threshold, which is the equivalent of spiking for non-spiking compartments. By using the Loihi's "PASS" operation to process CI's inputs, CI integrated the voltage from Inp only when AM's voltage was above threshold. This allowed the neuron to change its activity based on the sensory input only when the input was prominent enough to depolarize AM above its threshold, introducing a condition for the neural bursting. CI gave the bursting neuron's membrane voltage. As this Loihi compartment was non-spiking, our neuron model was lacking two crucial mechanisms: (i) spiking bursts and (ii) resetting after the bursts. To address the first issue, we added the somatic compartment (S) as the parent of CI, to receive from CI the information on the voltage exceeding the respective threshold. We used the Loihi's "OR" operation for S to spike whenever CI's voltage was above its threshold. For addressing the second issue (voltage reset), an inhibitory compartment (IN) was added. When IN fired, it decreased CI's voltage and reset the neuron to its resting state. By modifying the number of spikes from S that were necessary to make the IN compartment spike and reset CI, we obtained a bursting neuron with adjustable burst duration.

\subsection{Single Joint Control using Bursting Neurons}
We employed the bursting neuron as the building block for controlling each of the robotic joints. Specifically, the periodic bursting was used as a pacemaker for the repetitive flexion/extension of the joints. For a single tibia joint (Fig. 2B), we connected a bursting neuron to a pair of flexor-extensor motorneurons (FMN/EMN). We used the sensory input to drive the bursting neuron and this, in turn, excited the FMN and inhibited the EMN to form the flexion/extension alternation. We drove the EMN with tonic excitation, which normally comes from the higher brain centers \cite{soffe1982}. The spikes of the EMN/FMN were then used to drive the servomotors and move the joint. To do so, we decoded each spike of the FMN/EMNs to an increment of the joint's angle, as described by the following equation:
\begin{equation}
    \theta \xleftarrow{spike} \theta + \delta \theta, \qquad \delta \theta = \frac{\Delta \theta_{max}}{T_{j}},
\end{equation}
where $\theta$ is the joint's angle, $\delta \theta$ is its increment, $\Delta \theta_{max}$ is its maximum range, and $T_{j}$ is the tolerance of the joint to incoming spikes, i.e. the maximum number of spikes of a FMN/EMN that move the joint to its full range.

\subsection{Spiking CPG for locomotion}
To control the robot’s locomotion, we designed a CPG network, consisting of simple joint controllers (Fig. 2C). To achieve a tripod gait, in which at least three of the robot's legs were always in contact with the ground for stability, we grouped the robot's legs into two triplets, the right and the left. The right triplet comprised of the front and hind leg of the robot's right side and the middle leg of the robot's left side. The left triplet consisted of the remaining three legs. Each leg was controlled by two joints (Fig. 1), the tibia joint which lifted and lowered the leg, and the coxa joint which moved the leg back and forth. 

First, we needed to generate the alternating movement of the two triplets, referred to as the triplet cycle. For this, we connected two mutually inhibiting bursting neurons ($BN_{\mathcal{L}}$ and $BN_{\mathcal{R}}$), which fired in an alternating fashion and each of them controlled the movement of one triplet. Hence, the two triplets never moved simultaneously, which stabilized the robot on the ground. Preserving this alternating motion required the full cycle of a single leg (lifting-forward-lowering-backward) to be completed in half the duration of the triplet cycle. For this, we connected $BN_{\mathcal{L}}$ and $BN_{\mathcal{R}}$ to the joints of the respective limbs through an additional layer of bursting neurons, the tibia bursting neurons ($BN_{T_{i}}$). We designed the $BN_{T_{i}}$ to have a bursting frequency that is double the frequency of the triplet cycle. As a result, the $BN_{T_{i}}$ completed one full leg cycle before $BN_{\mathcal{L}}$ and $BN_{\mathcal{R}}$ alternated.   

To mobilize a single leg, we first connected the $BN_{T_{i}}$ to the tibia FMN/EMN ($F_{T_{i}}$/$E_{T_{i}}$), as described in section 2.2 (Fig. 2B). This allowed us to control the flexion/extension of these joints and, consequently the lifting/lowering of the legs. Tonic excitation drove the tibia EMNs ($E_{T_{i}}$) to reset the joints after the flexion. The back and forth leg movement was accomplished through the control of the coxa joints of the leg. The forward movement of a leg by its coxa flexion coincided temporally with both, the lifting of the same leg by its tibia flexion, and the backward movement of the contralateral leg by its coxa extension. Hence, we connected the $BN_{T_{i}}$ to the coxa FMN ($F_{C_{i}}$) of the same leg and the coxa EMN of the contralateral leg ($E_{C_{i+3}}$).To prevent the tibia joint extensions from interfering with their opposite leg cycles, even under persistent tonic excitation, we included contralateral inhibitory connections. These connections allowed the tibia FMN/EMN pairs ($F_{T_{i}}$/$E_{T_{i}}$) to suppress the tibia EMNs on the opposite leg ($E_{T_{i+/-3}}$). 

The spiking activity of the twelve pairs of FMNs/EMNs was decoded to joint angles to control the servomotors, as described in section 2.2.

\subsection{Loihi-ROS Interaction for Neuromorphic Implementation}
To achieve online robotic control, we developed a faster-than-real-time interaction between the Loihi Kapoho-Bay USB chipset and ROS, and utilized it to control our hexapod robot (Fig. 1). Our SNN CPG controller utilized less than 10\% of resources of a single Loihi core (64 compartments and 68 connections). Given that Loihi is an asynchronous system, we expanded our Loihi Astrocytic Module \cite{tang2019} to synchronize the chip with the ROS. This was done by blocking the faster system and forcing it to wait for the slower one. The ROS node iterated 100 times/s to sustain smooth control of the hexapod. For synchronization, $T_{epoch}$ Loihi iterations were executed during a single iteration of the robotic control loop. Each robotic control loop iteration included: i) transferring neuron spikes from Loihi to ROS, ii) decoding the spikes into servo positions (by increasing or decreasing the joint angles by a fixed amount $delta$ until the pre-defined limits - $max$ or $min$), and iii) setting servo positions through ROS topics. We set $T_{epoch}$ to 10, so that each Loihi iteration represented exactly 1 ms. The ``chokepoint'' was the communication between Loihi and the local host machine that ran the ROS node. To avoid data transfer in every Loihi iteration, we implemented a Sequential Neural Interacting Process (SNIP) on the x86 LMT chip (within the Loihi chipset) to store the number of spikes for $T_{epoch}$ iterations. 

\begin{figure}[!b]
\vspace{-15.2pt}
\centering
\includegraphics[scale=1.0]{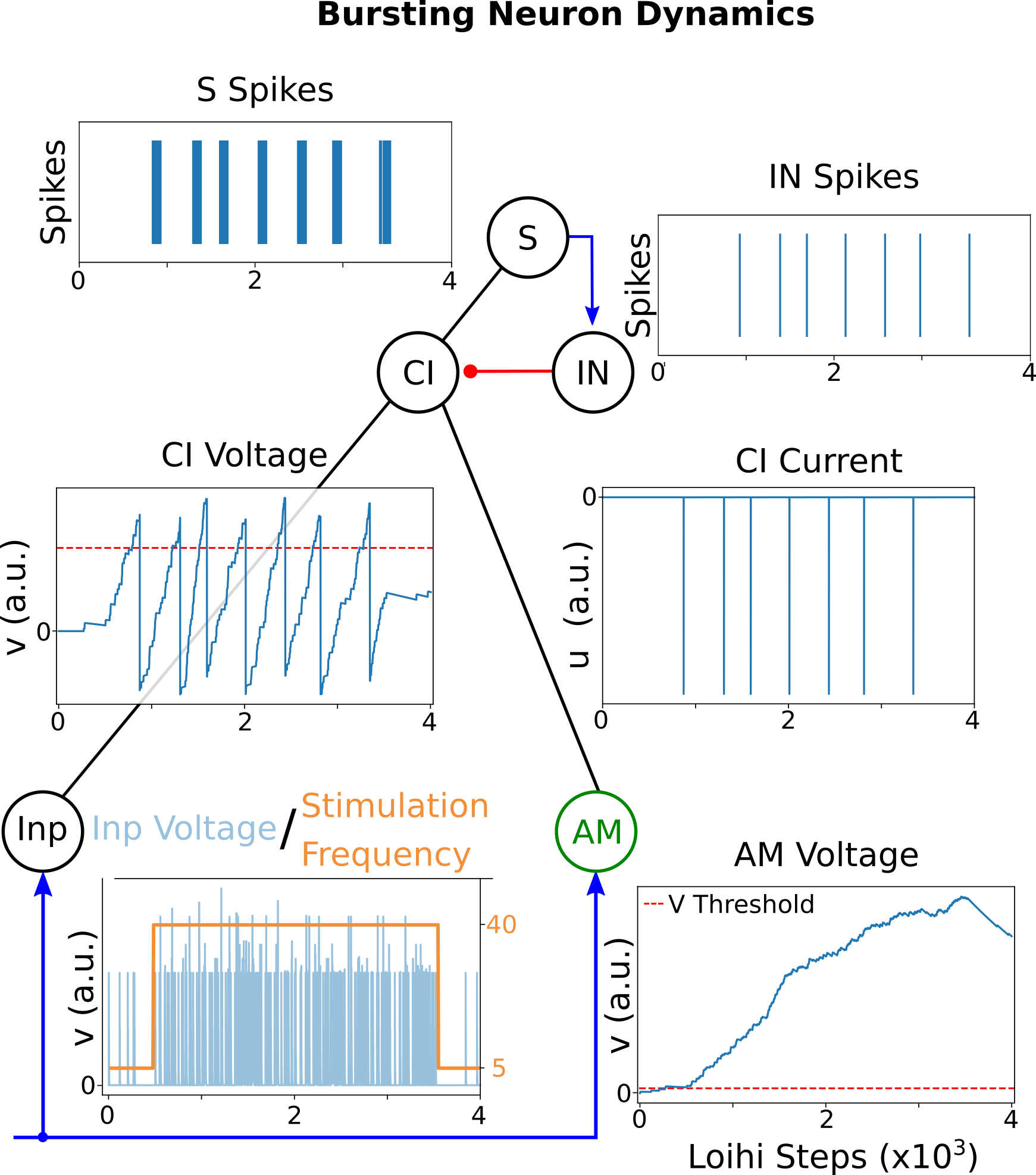}
\caption{Activity of the bursting neuron's compartments. Inp voltage is a direct transformation of the input current to voltage. AM is the integral of the input current. CI's voltage is the bursting neuron's membrane voltage; when it exceeds the threshold (red dotted) the somatic compartment S spikes. These somatic spikes drive IN to spike and stimulate CI to reset the neuron.}
\label{fig: BN}
\vspace{-5.2pt}
\end{figure}

\section{Results}
To test whether, and to what extent, the expected behavior can emerge out of the CPG network, we observed the kinematics of the robot’s joints and the robot's motion with respect to the sensory input. Here, we first report the internal dynamics of the bursting neuron, and then we relate this cellular behavior to that of the CPG network in controlling the robot’s gait cycle. We then examine the relation between the speed profile of the robot and the sensory input's frequency. Finally, we showcase the real-time interaction of Loihi with ROS and our hexapod robot.

\subsection{A Bursting Neuron Responding Only to Meaningful Stimuli}
The activity within the bursting neuron's compartments is shown in Fig. \ref{fig: BN}. In this example, we stimulated the neuron with a 4s Poisson spike train. For the first 500 ms, the firing frequency was kept to its baseline 5Hz, representing sensory noise. Between 500ms and 3500ms, the firing frequency increased to 40Hz, representing sensory neuron activity that encodes input stimulus. Using the mechanisms described in section 2.1, Inp transformed its input current to voltage (Fig. 3, bottom-left panel), and AM integrated the input to its voltage (Fig. 3, bottom-right panel). This allowed for the sensory noise to be filtered out, as it was only when the frequency increased to 40 Hz that CI started integrating the voltage of its child node, Inp, into its own (Fig.3, middle-left panel). The S compartment received from CI the information about whether CI's voltage exceeded its threshold. The S compartment continued spiking (Fig. 3, top-left panel) for as long as CI's voltage was above its threshold. This created the spike burst activity. In turn, the spikes of the S compartment stimulated IN, which integrated them in its voltage. The spikes of IN (Fig. 3, top-right panel) generated inhibitory currents that closed the loop by being fed into CI (Fig. 3, middle-right panel). This current decreased CI's voltage (Fig. 3, middle-left panel) and reset the neuron to the resting state. The time needed for CI's voltage to reach the threshold again, after it was reset, was the inactive period between spike bursts.

\begin{figure}[!t]
\vspace{-5.2pt}
\centering
\includegraphics[scale=1.06]{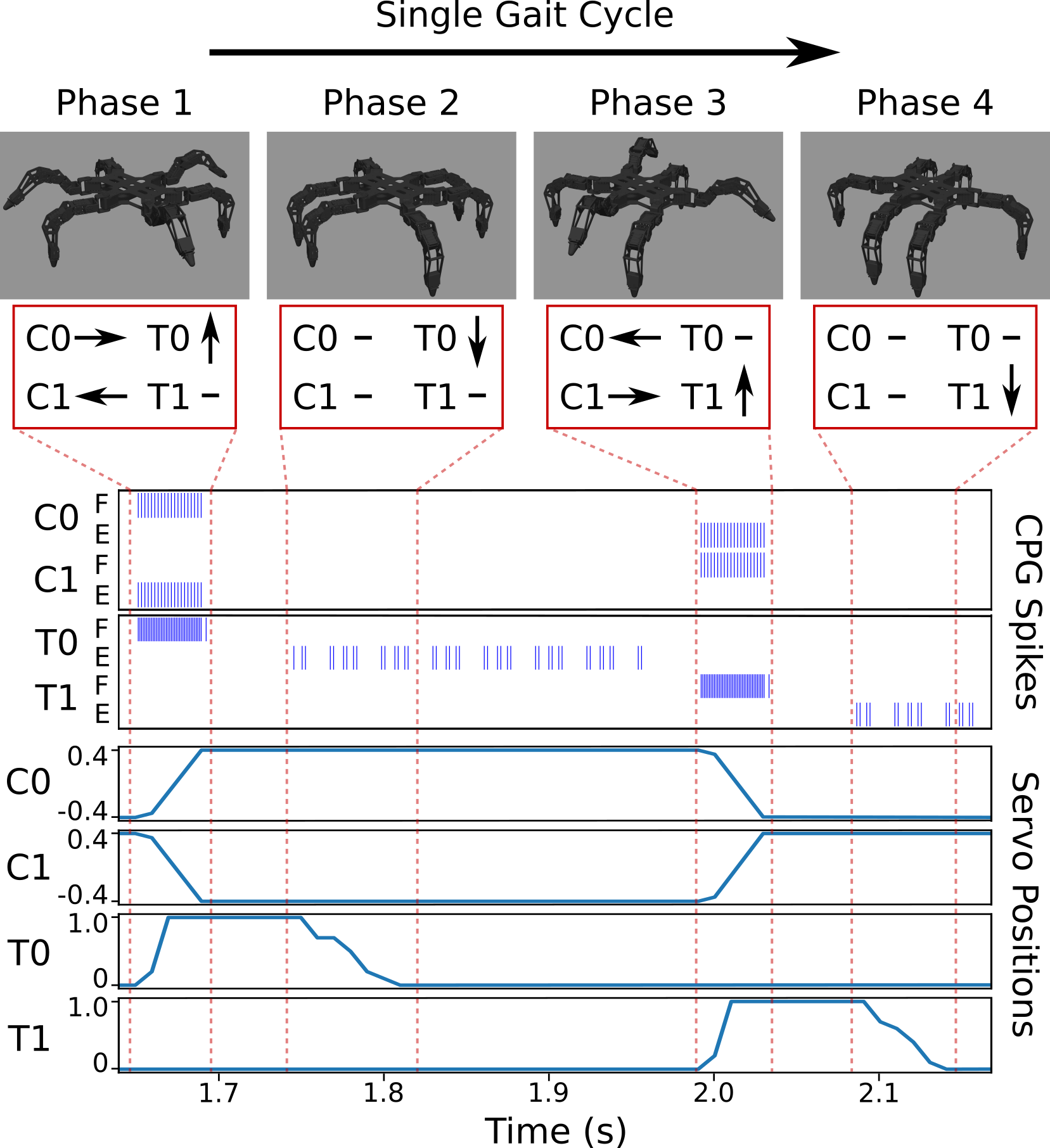}
\caption{Example of a single hexapod gait cycle controlled by the spiking CPG. Raster plot for motor-neuron spiking activity in the CPG resulted in sequential change of the servos' position, for both flexion and extension.}
\label{fig: gait}
\vspace{-15.2pt}
\end{figure}

\subsection{Gait Cycle}
To achieve the regular walking pattern of the robot, a single gait cycle (Fig. 4) was repeated, as described below. During the gait cycle, the BN of the tibia ($BN_{T0}$) drove both the coxa FMN of the same leg ($F_{C0}$) and the coxa EMN of the opposite leg ($E_{C1}$). In this way, a leg was lifted by its tibia joint, while its coxa joint moved it forward. Simultaneously, the coxa joint of the opposite leg moved it backwards (Phase 1). Then, the extension of the tibia joint lowered the leg to the ground (Phase 2). This concluded half of the gait cycle and was followed up by the other half (Phases 3 and 4). During this next half, the coxa joints of the legs that had previously moved forward ($C_{0}$) now moved them backwards and vice versa ($C_{1}$). Likewise, the legs that were previously in contact with the ground were now lifted by their respective tibia joints ($T_{1}$). 

\begin{figure}[!b]
\vspace{-15.2pt}
\centering
\includegraphics[scale=1.06]{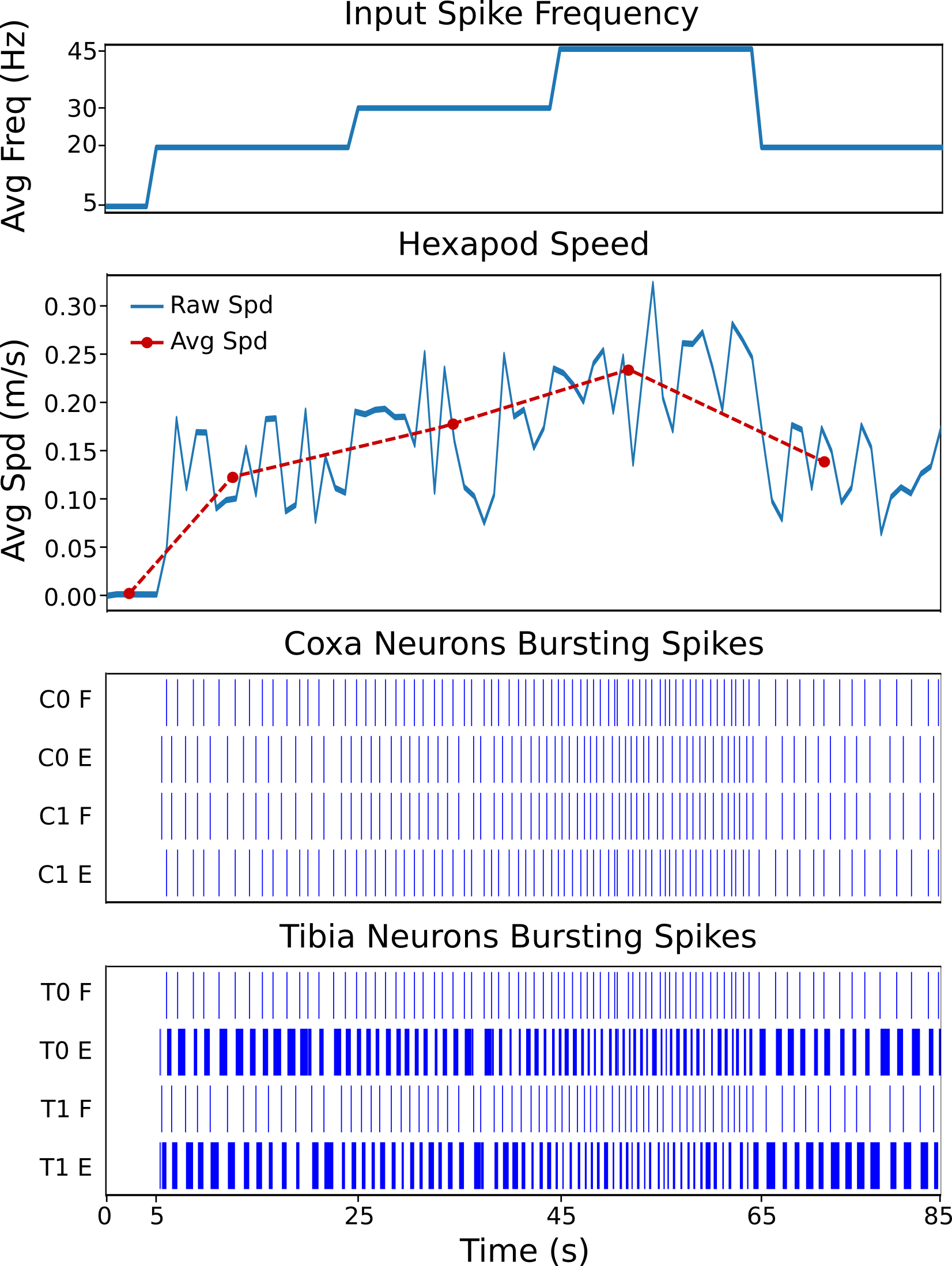}
\caption{Acceleration and deceleration of the hexapod by changing the input firing activity to the CPG every 20 seconds in a single experiment.}
\label{fig: dynmic_spd}
\vspace{-5.2pt}
\end{figure}

\subsection{Control of the Robot's Speed Using the Input frequency}
The dependence of the robot's speed on the input frequency emerged out of our CPG design. We used random Poisson spike trains with a mean firing rate to test the ability of our CPG network to generate a periodic output given a non-periodic input. We chose 4 representative input frequencies to stimulate the CPG network and tracked the robot's speed. Specifically, we first drove our CPG network with a 5Hz Poisson train that simulated sensory noise, as in the case of the single BN (section 3.1). As expected, this baseline sensory activity was insufficient to depolarize the AM compartment of the $BN_{\mathcal{L}}$ and $BN_{\mathcal{R}}$, and consequently, could not generate any robot motion. We then increased the sensory input frequency beyond the noise level. When the firing activity exceeded the threshold, the AM compartment started getting depolarized and the CPG network generated the motion pattern based on the gait cycle described in section 3.2. In Fig. 5, we show the speed of the walking robot (second panel) in response to its input frequency (top panel) which increased from 5Hz to 20Hz (for t=[0,20]s), 30Hz (for t=[20,40]s) and 45Hz (for t=[40,60]s). This resulted in a smooth increase of the robot’s average speed (Fig. 5, second panel; red dotted line). The spiking frequency of the joints' FMNs/EMNs successfully adapted to the input frequency, as shown in the raster plots (Fig. 5, two bottom panels.) 

\subsection{Speed Profile in Response to Stimulus Frequency}
We tracked the robot's speed as a function of input frequency over the whole functional range of our CPG network. For this, we stimulated the CPG network with 60s Poisson trains, whose frequency ranged between 20Hz and 50Hz with a step of 5Hz. This was done 10 times for each input frequency, and we recorded the robot's average speed for each such trial. Then we computed the mean of these average speeds to obtain the average speed profile, shown in Fig. 6. This plot describes the effective range of the movement speed as a function of the input sensory frequency.

\subsection{Loihi-ROS Interaction for a Real-Time Control of Robots}
Our framework exhibited a faster-than-real-time interaction between ROS and Loihi. To verify this, we measured the real-time factor (RTF) defined as follows:
\begin{equation}
    RTF = \frac{t_{exec}}{t_{wc}}
\end{equation}
where $t_{exec}$ is the execution time of a single iteration and $t_{wc}$ is the wall-clock time, which for a ROS node iterating at 100 Hz is always equal to 0.01s.  
When averaged over all the iterations, we obtained an RTF of 1.6, which is far above the minimum requirement of a real-time system (Fig. 7).
Consequently, our ROS-Loihi interaction framework can indeed be used to control a real-time hexapod robotic system. In fact, since this framework is not hardware specific, it can be used to control other robotic systems as well. 

Moreover, since $T_{epoch}$ Loihi iterations were executed during a single iteration of the robotic control loop, this reduced the communication cost between Loihi and the local host machine by a factor of $T_{epoch}$.

\begin{figure}[!b]
\vspace{-5.2pt}
\centering
\includegraphics[scale=1.23]{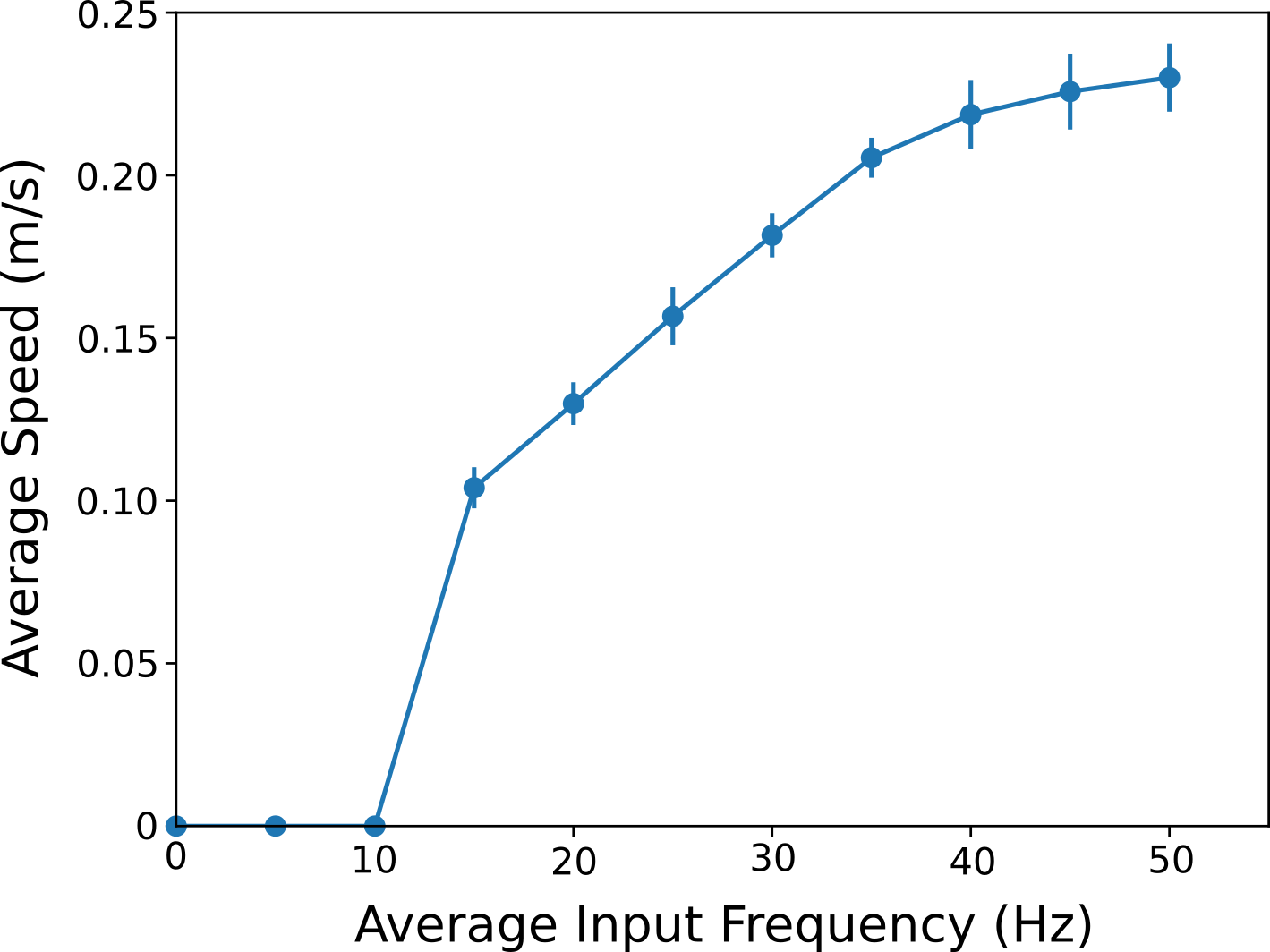}
\caption{Average speed of the hexapod for different input frequencies. Each data point is the mean of 10 average speed recordings obtained over 60s long experiments. The vertical line shows the speed range within the trials.}
\label{fig: mean_spd}
\vspace{-5.2pt}
\end{figure}

\section{Discussion}
In this paper, we presented a comprehensive neuronal-astrocytic CPG network and its integration into Intel’s Loihi neuromorphic processor, to control the locomotion of a hexapod robot. We also proposed a framework for the real-time interaction between the ROS environment and Loihi. This work aims to translate the advantages, including that of simplicity, carried by traditional oscillator-based CPG models \cite{buono2001, sartoretti2018} into the emerging neuromorphic chips that promise significant energy-efficiency, robustness and versatility in solving real-world robotic problems. Unlike conventional approaches that use dynamical systems as coupled oscillators \cite{ijspeert2008,crespi2006,buono2001,sartoretti2018}, our proposed neuromorphic control algorithm generated the periodic movement pattern by virtue of the modeled brain cell's properties, as well as the interaction between neuronal and astrocytic activity. 

\begin{figure}[!t]
\vspace{-5.2pt}
\centering
\includegraphics[scale=1.14]{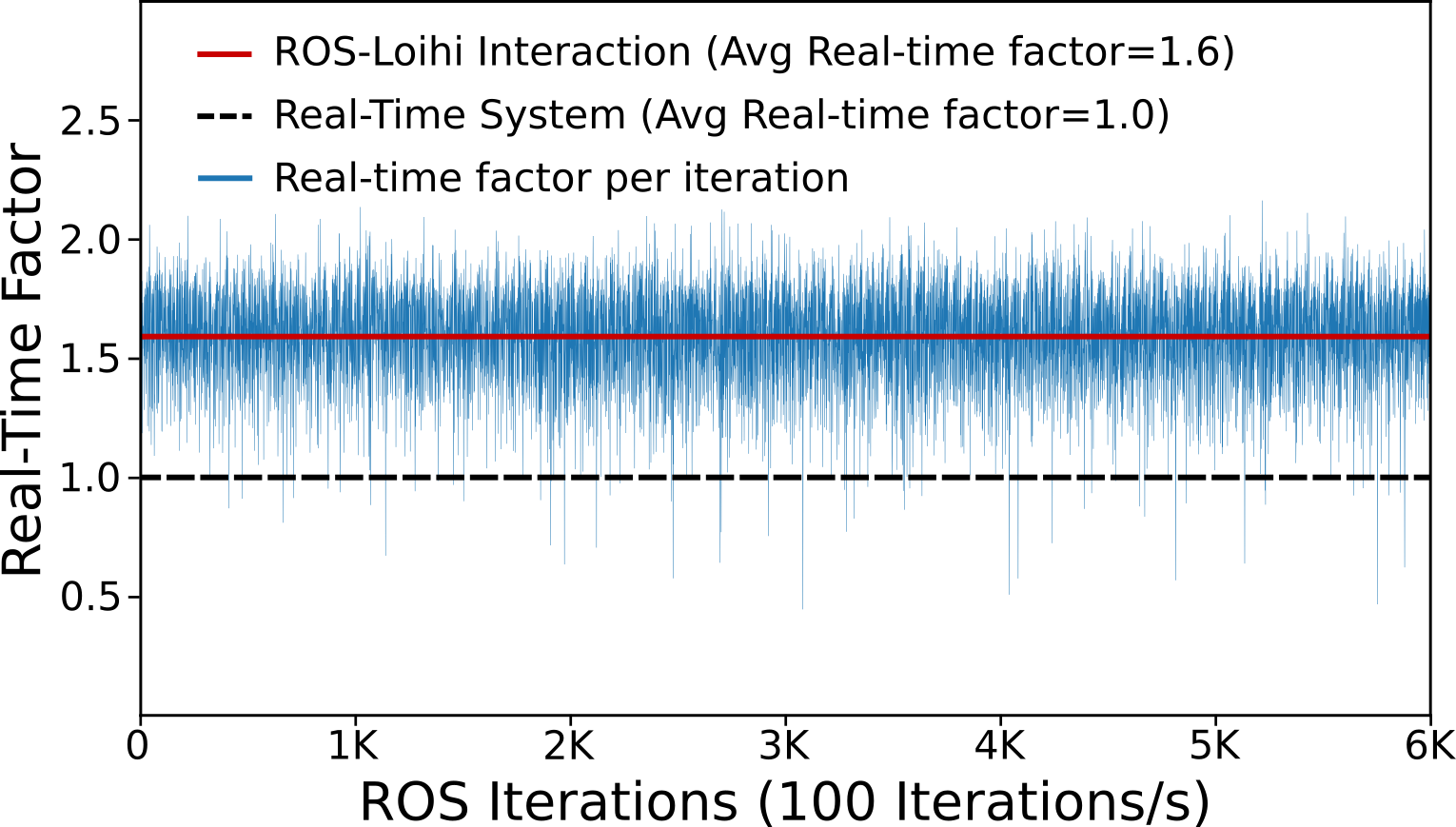}
\caption{The real-time factor for the spiking CPG system based on time recordings from each ROS iteration.}
\label{fig: realtime}
\vspace{-15.2pt}
\end{figure}

At the cellular level, we introduced the first conditionally-bursting neuron on a neuromorphic hardware drawing inspiration from biological findings in insect locomotion \cite{grillner1975, mccrea2007, delcomyn1999}. Current neuromorphic chips provide an efficient hardware implementation of the LIF neuron model only, and, therefore, lack an inherent mechanism for emulating the bursting neuronal activity that is present in the brain \cite{morquette2015,mccrea2007}. While LIF neurons are far from ideal when driving periodic behavior, our proposed CPG model is a perfect fit for a behavior that repeated itself every few hundred milliseconds -- a time window that coincides with the behavioral time scale. Such long periodic interchanges between active and inactive periods are primed for tasks where robustness to sensory neuronal noise matters. The proposed bursting neurons encode the oscillatory activity in a bandpass manner (using their inter- and intra-burst bands), which makes them robust to noise present in the full operating spectrum. Our approach also draws from the recent impressive empirical evidence of the neuromodulatory abilities that astrocytes exhibit \cite{morquette2015}. Indeed, the astrocyte added a computational layer to the brain-morphic CPG, which was orthogonal to the neuronal layer, and enabled switching of the network's behavior depending on the input stimulus. The astrocytic component decreased the sensitivity of our network to low-frequency sensory activity, which represents noise in the adopted frequency-encoding regime. In contrast to the fast spiking noise-prone neurons, astrocytes can use their much larger temporal and spatial integration scales \cite{polykretis2018, polykretis2019} to filter out noise transients from meaningful stimuli. We expect this approach to become a significant component in our and others' efforts towards robust spiking neural networks. In that sense, this work captures the newly identified roles of astrocytes and solidify them as computational primitives applicable to spiking neuronal-astrocytic networks integrated to large-scale neuromorphic chips.

At the network level, the CPG connectome was inspired, and partially constrained, by its biological counterpart \cite{grillner1975, mccrea2007, delcomyn1999}. This significantly decreased the size of our proposed network, which consisted of only 64 Loihi compartments (less than 0.5\% of the 15K-compartment network in \cite{Guangzhi} that consumed just 9mW). Although we did not measure the power consumption of our network, the asynchronous computation on Loihi that consumes power proportionally to the utilized resources make it reasonable to expect significant power efficiency. The network's architecture also ensured a range of linear dependence, in which the period of the gait cycle was inversely proportional to the frequency of the sensory input. Contrary to typical neuron models in conventional multi-layer networks that can be optimized to perform complex computational tasks, spiking neuron models have a non-differentiable output (their all-or-none firing) and therefore are incompatible with standard gradient-descent supervised learning methods. In the absence of a strong learning algorithm, the main criticism to neuromorphic solutions is that promising preliminary results \cite{Guangzhi} cannot share the same scaling abilities with the mainstream deep learning approaches. Adding biological constraints to neuromorphic algorithms, such as the ones we introduced here, removes the need for assuming all-to-all initial connectivity for the trainable network. This may translate to further improvements in training efficiency, as it limits learning to a small number of synaptic connections. Recent applications of gradient-descent alternatives to SNNs \cite{RN255, RN28} are also promising but they inherit the main limitations that the conventional neuronal networks have. For example, learning through backpropagation will basically match the network’s input to its output, much regarding and, thereby, structuring the network as a black-box. Bottom-up neuromorphic approaches in developing neural-controlled robots, such as the one proposed here, can give an additional benefit, by serving as test-beds to inform brain scientists on how the neural system could be structured to function properly.  In that sense, this work contributes to a novel direction, where ``machine behavior'' in general, and robotic function in particular, can emerge naturally from a basic \textit{a-priori} knowledge of its controller’s structure. 

The coupling between the robot’s speed and the sensory input frequency is crucial in our CPG design as it may give rise to a number of distinct control strategies that depend on the sensory input. Consider, for example, a visual stimulus being the input to the proposed CPG; The representation of the distance from a desirable target using the input frequency (rate encoding) would lead to a robotic speed that depends on how close the robot is to the perceived target: A large distance would be encoded as high input frequency and cause the robot to move fast towards its target; As the robot approaches the target, the speed decreases. In contrast, encoding the distance from a dangerous object into the input frequency could force the robot to move fast away from that object, when being in close proximity to it. Therefore, our proposed CPG design allows for multiple parallel perception-action loops, that drive the same spiking network and can modify the robot's speed in response to a number of sensed changes in a varying environment. It is true that the embodiment of spiking neural networks into robots has been rather sparse and most of the current approaches aim to give a proof of concept \cite{RN270, RN278}, rather than a whole-behaving robot. While there is definitely value in studying simplified tasks and basic sensory representations \cite{RN279}, there is an ongoing need to propose new architectures capable of naturally handling richer, noisier and more complex scenarios \cite{RN218}. Our work paves the way for developing novel active sensing neuromorphic solutions by drawing biological principles from neuroscience and translating them to computational primitives.

When it comes to innervating robots, there is no scarcity of communication frameworks between ROS and most of the popular neuron simulators \cite{weidel2016closed}. Currently, however, there is no coupling between the most recently introduced neuromorphic chip, Loihi, and ROS. To propose an end-to-end solution, we had to bridge the gap between our on-chip CPG network and the robot's joints, by proposing a real-time intercommunication framework. For a reasonable communication to take place, the control commands need to be generated before they are required by the robot. Our framework established a faster-than-real-time intercommunication between ROS and Loihi that can further incorporate more computationally intensive Loihi networks that would communicate with ROS and add up to the functionality of the robot. 

Overall, an autonomous robot should, among others, 1) be robust to a noisy neural representation, 2) adapt to a fast changing environment, and 3) learn with no or limited supervision or reinforcement. The emergence of neuromorphic computing calls for a bottom-up rethinking of the real-time control algorithms for robots, that can seamlessly integrate into non-Von Neumann hardware, promising unparalleled energy-efficiency and a robust yet versatile alternative to the brittle inference-based AI solutions. This paper brings us closer to realize this promise, focusing on the CPG, a crucial mechanism for legged robots.

\section{Conclusion}
Effective as they may have become, robots still cannot duplicate a range of human behaviors, such as responding to changing environments using error-prone sensors. This work draws from the structure of the brain areas associated with the targeted behavior as well as the recently identified principles for information processing in the brain and introduces a neuromorphic approach in designing robotic controllers. The reported robotic behavior was achieved by emulating the connectome and the underlying types of brain cells, not through learning. This suggests that the co-development of brain-morphic algorithms and neuromorphic hardware for solving scalable robotic problems is a direction worth pursuing.


\bibliographystyle{ACM-Reference-Format}
\bibliography{sample-base}

\end{document}